\documentclass{article}

    \PassOptionsToPackage{numbers, compress}{natbib}

 \pdfoutput=1

   \usepackage[nonatbib, final]{neurips_2022}

\usepackage[utf8]{inputenc} %
\usepackage[T1]{fontenc}    %
\usepackage{hyperref}       %
\usepackage{url}            %
\usepackage{booktabs}       %
\usepackage{amsfonts}       %
\usepackage{nicefrac}       %
\usepackage{microtype}      %
\usepackage{xcolor}         %
\usepackage{graphicx} 

\usepackage{subcaption}
\usepackage{xcolor}
\usepackage{colortbl}

\usepackage{makecell}
\usepackage{pdfpages}
\usepackage{enumitem}
\usepackage{gensymb}
\usepackage{multirow}
\usepackage{amsmath,amssymb}
\usepackage{wrapfig, lipsum}
\usepackage{ragged2e}
\usepackage{pifont}%
\newcommand{\cmark}{\ding{51}}%
\newcommand{\xmark}{\ding{55}}%
\usepackage{xspace}

\newcommand{\etal}{\textit{et al}.}

\DeclareMathOperator*{\prompt}{\text{Prompt}}

\DeclareSymbolFont{largesymbolsCM}{OMX}{cmex}{m}{n}   
\DeclareMathSymbol{\sumop}{\mathop}{largesymbolsCM}{"50}

\def\ours{\texttt{\textbf{DualCoOp}}\xspace}

\title{DualCoOp: Fast Adaptation to Multi-Label Recognition with Limited Annotations}

\author{Ximeng Sun$^{1}$ \ \ \ \ Ping Hu$^{1}$ \ \ \ \ Kate Saenko$^{1,2}$ \\
$^{1}$Boston University, $^{2}$MIT-IBM Watson AI Lab, IBM Research \\
{\tt \small \{sunxm, pinghu, saenko\}@bu.edu}}

\begin{document}

\maketitle

\begin{abstract}
Solving multi-label recognition (MLR) for images in the low-label regime is a challenging task with many real-world applications.
Recent work learns an alignment between textual and visual spaces to compensate for insufficient image labels, but loses accuracy because of the limited amount of available MLR annotations. 
In this work, we utilize the strong alignment of textual and visual features pretrained with millions of auxiliary image-text pairs and propose \textit{Dual Context Optimization} (\ours)  as a unified 
framework for partial-label MLR and zero-shot MLR. \ours encodes positive and negative contexts with class names  as part of the linguistic input (i.e. prompts). Since \ours only introduces a very light learnable overhead upon the pretrained vision-language framework, it can quickly adapt to  multi-label recognition tasks that have limited annotations and even unseen classes.  Experiments on  standard multi-label recognition benchmarks across two challenging low-label settings demonstrate the advantages of our approach over state-of-the-art methods. Our code will be publicly available.

\end{abstract}

\section{Introduction}
Image recognition has become a very popular and successful research area in recent years, due to the development of large-scale datasets~\cite{deng2009imagenet,kuznetsova2020open} and advanced model architectures~\cite{dosovitskiy2020image,he2016deep,liu2021swin,simonyan2014very}. 
However, the majority of image recognition approaches have focused on single-label prediction, which ignores the intrinsic multi-label nature of images.
Unlike single-label recognition~\cite{dosovitskiy2020image,he2016deep,liu2021swin,simonyan2014very}, multi-label image recognition aims to recognize all semantic labels present in an image~\cite{chen2019multi,chua2009nus,liu2017semantic,liu2018multi,sarafianos2018deep,yazici2020orderless,wang2020multi}, providing a more comprehensive understanding and benefiting applications like image retrieval, video analysis, and recommendation systems.

Multi-label recognition typically deals with images of complex scenes and diverse objects. Collecting multi-label annotations becomes difficult to scale up, for two reasons: (i) annotating images with the full semantic label set is laborious and (ii) samples of particular categories can be hard to find. 
The first challenge can be addressed by multi-label recognition with \textit{partial labels}, where merely some of the categories are annotated for each training image. 
Recent works 
proposed solutions to partial-label MLR based on semi-supervised learning~\cite{joulin2016learning,mahajan2018exploring}, normalized training objectives~\cite{durand2019learning}, or label correlations~\cite{chen2022structured,huynh2020interactive,pu2022semantic}.
The second setting involves \textit{zero-shot} MLR, where novel unseen categories are recognized by transferring knowledge from seen categories, with solutions like principal image features~\cite{ben2021semantic,zhang2016fast}, knowledge graphs~\cite{lee2018multi}, and attention mechanisms~\cite{huynh2020shared,narayan2021discriminative}.
Despite significant progress on the two settings, existing approaches are not designed to handle both. We propose to unify these settings as \textit{limited-annotation} MLR and design a solution that can handle practical scenarios with either partial or missing labels.

Successful solutions to the above problems transfer knowledge from fully-annotated categories to partially-labeled and novel categories by learning an alignment between images and category names~\cite{chen2022structured,pu2022semantic,zhang2016fast}. 
Recently, vision-language pretraining models are bridging the visual-textual gap via large-scale pretraining, e.g., CLIP~\cite{radford2021learning} is trained with 400 million image-text pairs. In this work, we draw inspiration from the recent success of prompt learning for such models~\cite{huang2022unsupervised,ju2021prompting,luddecke2021mm,radford2019language,zhou2021denseclip,zhou2022conditional}. 
Prompt learning provides a convenient way to transfer pretrained vision-language models to other tasks. 
It designs additional templated or learnable prompt tokens for textual input to “inform” the model about downstream tasks and avoids finetuning the entire model which can be inefficient and data-hungry. 
By doing so, recent works like CoOp~\cite{zhou2021learning} have demonstrated CLIP's remarkable generalisation to various zero-shot image tasks~\cite{huang2022unsupervised,radford2021learning,zhou2022conditional}.
However, these methods mainly focus on matching each image with a single label, hence they are not able to handle the multi-label setting.

\begin{figure}[t]
\begin{center}
     \includegraphics[width=0.85\linewidth]{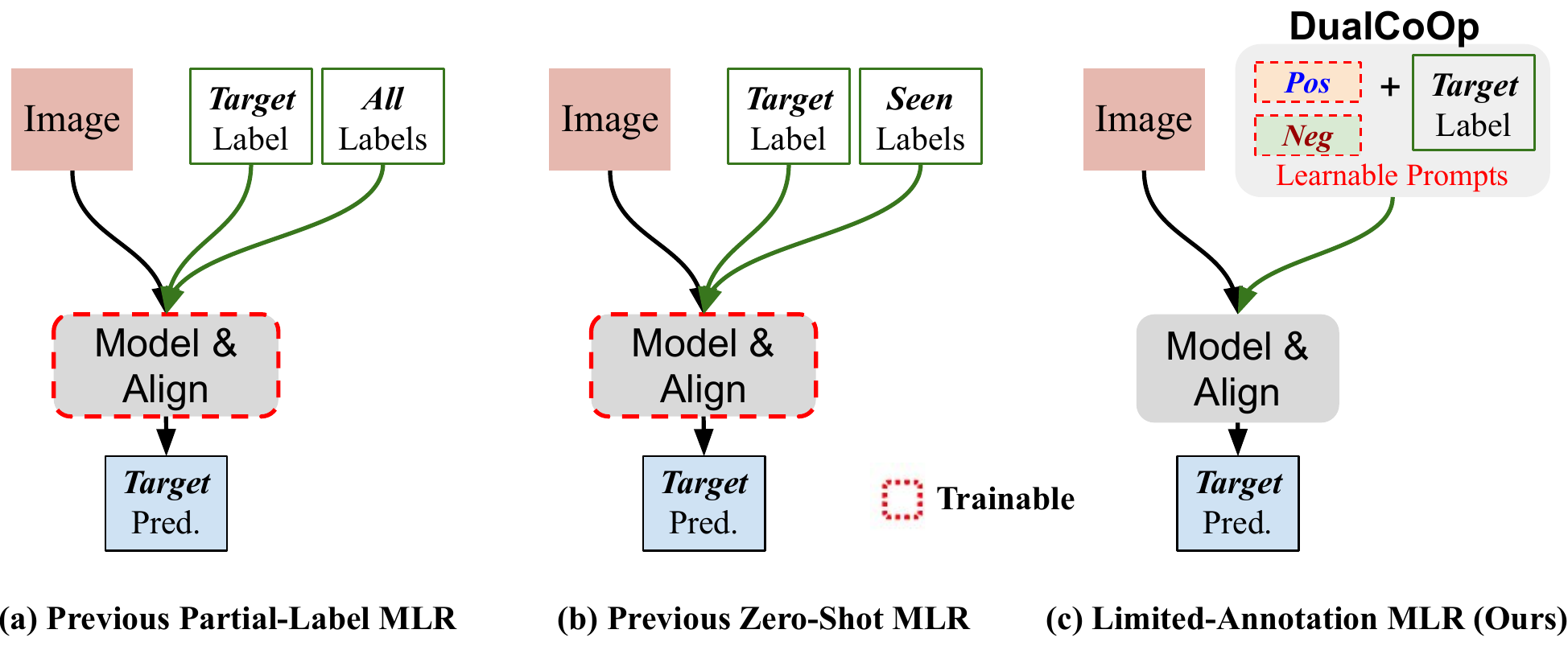}
\end{center} %
   \caption{\small 
   \textbf{A conceptual comparison of previous multi-label recognition (MLR) methods and our approach}. In Part-Label MLR and Zero-Shot MLR, previous works learn to model and align the visual and textual inputs as well as explore the correlation between the target label with all/seen labels depending on the limited semantic annotations available on the dataset, which leads to sub-optimal performance and complex model designs. In contrast, we propose a unified framework to tackle both tasks with limited annotations. We learn a pair of positive and negative prompts as the textual input and rely on the modeling and alignment of visual and textual inputs fully on the precise alignment of vision-language pretraining on a large-scale dataset. 
   }
   \vspace{-10pt}
   \label{fig:concept}
\end{figure}

To adapt the knowledge learned in CLIP to multi-label image recognition, we propose the \ours framework. 
As shown in Fig.~\ref{fig:concept} (c), \ours learns a pair of differentiable prompts to provide positive and negative contexts for the target class. 
Instead of using hand-crafted thresholding to determine positive labels~\cite{ridnik2021asymmetric},  the dual prompts naturally result in a positive and a negative classifier, so the existence of the target class in the image can be easily decided by comparing their scores.
Unlike prior models, shown in Fig.~\ref{fig:concept} (a)(b), we avoid 
fine-tuning the full vision-language model and only learn the prompts, which are much smaller compared to the entire model. Therefore,  our simple framework achieves much higher efficiency when adapting to different datasets.
Additionally, we modify the attention mechanism of CLIP to better model spatial information in images, improving its ability to recognize multiple objects in MLR.
With these design choices, we achieve a unified framework for addressing the general challenges of multi-label recognition with limited annotations. 

We summarize our contributions as follows:
\begin{itemize} [leftmargin=*]
    \item We propose \ours to quickly adapt powerful vision-language models to solve multi-label recognition tasks using limited annotations.
    \item We propose dual (positive and negative) prompts to drastically reduce the number of learnable parameters, and improve the spatial modeling of the visual encoder to better distinguish multiple objects.
    \item We conduct extensive experiments on  partial-label MLR (on MS-COCO~\cite{lin2014microsoft} and VOC2007~\cite{everingham2010pascal}) and zero-shot MLR (on MS-COCO and NUS-WIDE~\cite{chua2009nus}). Notably, \ours improves mAP by $6.8\%$ with $10\%$ of labels on VOC2007, and F1-score at Top-3 Prediction by $10.8\%$ for zero-shot MLR on NUS-WIDE. %
\end{itemize}

\section{Related Works}
\vspace{-10pt}
\vspace{1mm}
\textbf{Multi-Label Recognition with Limited Annotations.}
Multi-label image recognition has drawn increasing attention in past years. 
One straightforward solution to this problem is to individually learn a binary classifier for each category ~\cite{liu2015optimality,misra2016seeing,tsoumakas2007multi}, which however does
not consider correlations among labels. 
Hence, recent works have focused on incorporating semantic dependencies among labels via  graph neural networks~\cite{chen2019multi,chua2009nus,wang2020multi} or RNN/LSTM~\cite{liu2017semantic,wang2016cnn,wang2017multi,yazici2020orderless}. 
Some work also considers the spatial distribution of labels in the image, and exploits object proposals~\cite{li2016human,liu2018multi,wang2016beyond} or attention mechanism~\cite{sarafianos2018deep,wang2017multi,zhu2017learning} as a regularization to rectify the prediction.
However, despite achieving significant progress, these methods require a large-scale and complete annotated dataset to train models~\cite{krishna2017visual,lin2014microsoft}. 
This limits their application to more practical scenarios where data is partially annotated for training~\cite{bucak2011multi,chen2013fast,mahajan2018exploring,sun2017revisiting,sun2010multi,xie2018partial} and unseen (zero-shot) categories may appear during testing~\cite{chen2020knowledge,gupta2021generative,lee2018multi,mensink2014costa,zhang2016fast}. 

With partially labeled data, where merely some labels of each sample are known, Mahajan~\etal~\cite{mahajan2018exploring} and Joulin~\etal~\cite{joulin2016learning} attempt to use web supervision to automatically generate the pseudo labels, which unfortunately leads to poor performance as the web supervision is noisy and incomplete~\cite{zhang2021understanding}. To avoid external noise, Durand~\etal~\cite{durand2019learning} exploit the proportion of annotated samples for different labels and propose a normalized BCE loss to train models based on the given partial labels. 
More recent works explicitly transfer information from known
labels to complement unknown labels by utilizing  category-specific feature blending~\cite{pu2022semantic} or label co-occurrences~\cite{chen2022structured} at both instance-level and prototype-level.

Unlike partial annotation of the same label set for training and testing, zero-shot multi-label image recognition needs to handle novel categories during testing, hence inspiring a different route based on a joint visual-label embedding space~\cite{chen2020knowledge,huynh2020shared,mensink2014costa,narayan2021discriminative,zhang2016fast}. 
Zhang~\etal~\cite{zhang2016fast}
propose to find a  principal direction that ranks related labels first in the joint embedding space optimized via a tailored zero-shot ranking loss. Cohen~\etal~\cite{ben2021semantic} further improve the idea by learning multiple principal vectors to support the semantic diversity. 
Huynh~\etal~\cite{huynh2020shared} consider the spatial regularization and propose a shared multi-attention model and obviate the need for explicit region proposals~\cite{rahman2019deep0tag}. 
Narayan~\etal~\cite{narayan2021discriminative} propose to enhance the region-based features  so as to  minimize inter-class feature entanglement.

Though significant progress has been made in each of the directions, existing methods still require a lot of MLR data and complex architectures/losses. Our approach reduces the need for hard-to-get MLR data by pretraining on unsupervised text-image pairs. While it may seem unfair to compare existing MLR methods with ones based on such pretraining, we point out that the pretraining data is unsupervised and thus easier to obtain. We also provide experiments comparing \ours to baselines using the same
pretraining.
Importantly, previous methods are designed for only one task, hence have limitations in practical applications. In contrast, our proposed framework can be easily adapted with small data and can address both partial and zero-shot tasks at the same time.   

\vspace{1mm}
\textbf{Prompt Learning for Vision-Language Models.}
Vision-Language Models~\cite{jia2021scaling,radford2021learning} based on contrastive learning have demonstrated impressive ability to learn generic visual representations. As a milestone, CLIP~\cite{radford2021learning} is trained with 400 million curated image-text pairs, and shows remarkable transfer capability for over 30 classification datasets.
With such powerful vision-language models, several follow-ups~\cite{gao2021clip,wortsman2021robust,yao2021cpt,zhang2021tip} have been proposed to explore the training strategies for training downstream classification tasks.
Instead of fine-tuning the entire model~\cite{dong2019unified,he2016deep}, which may damage the learned representation space, recent approaches adopt  the prompt-based paradigm that formalizes NLP tasks as masked language modeling (prompt templates)~\cite{lester2021power,li2021prefix,shin2020autoprompt}.
Zhou~\etal~\cite{zhou2021learning} propose to tune prompts for downstream classification tasks, and further introduce input-conditional prompts for better generalization ability~\cite{zhou2022conditional}. Lu~\etal~\cite{lu2022prompt} learn the distribution of diverse prompts to handle the varying visual representations. Huang~\etal~\cite{huang2022unsupervised} generate pseudo labels for images to learn prompts in an unsupervised way.
Though achieving promising improvements for downstream tasks, these methods address the multi-class zero-shot image recognition, assuming each image has one label, hence lacking the ability to handle the multi-label setting.  
In this paper, we present a novel framework to efficiently transfer VLMs to address multi-label image recognition with limited annotations.

\section{Method}
\vspace{1mm}
\textbf{Problem Definition.}
We formally define multi-label recognition with limited annotations as follows: Consider $M$ as the set of categories which describe objects or attributes in images. Given a training image $I$, the existence of a category $m\in M$ can be positive, negative or  unknown, corresponding to the label $y_m = 1, -1$ or $0$ respectively. During inference, we predict each label of interest for an input image. 

Many existing MLR problems fit into this broad definition. In this paper, we consider the settings with partial or missing labels: (1) \textbf{Partial-label MLR}~\cite{chen2022structured,durand2019learning,pu2022semantic}, in which only a subset of labels are known ($+1$ or $-1$) for one training image and we are interested in predicting all existing labels during inference. (2) \textbf{Zero-shot MLR}~\cite{ben2021semantic, huynh2020shared, rahman2018deep}, in which each label is either known (seen) or unknown (unseen) for \textit{all} images during  training and we are interested in predicting either all labels or only unknown (unseen) labels during inference. 
In this paper, we propose a unified framework to address the \textit{limited-annotation} MLR in both scenarios.

\begin{figure}
\begin{center}
     \includegraphics[width=0.99\linewidth]{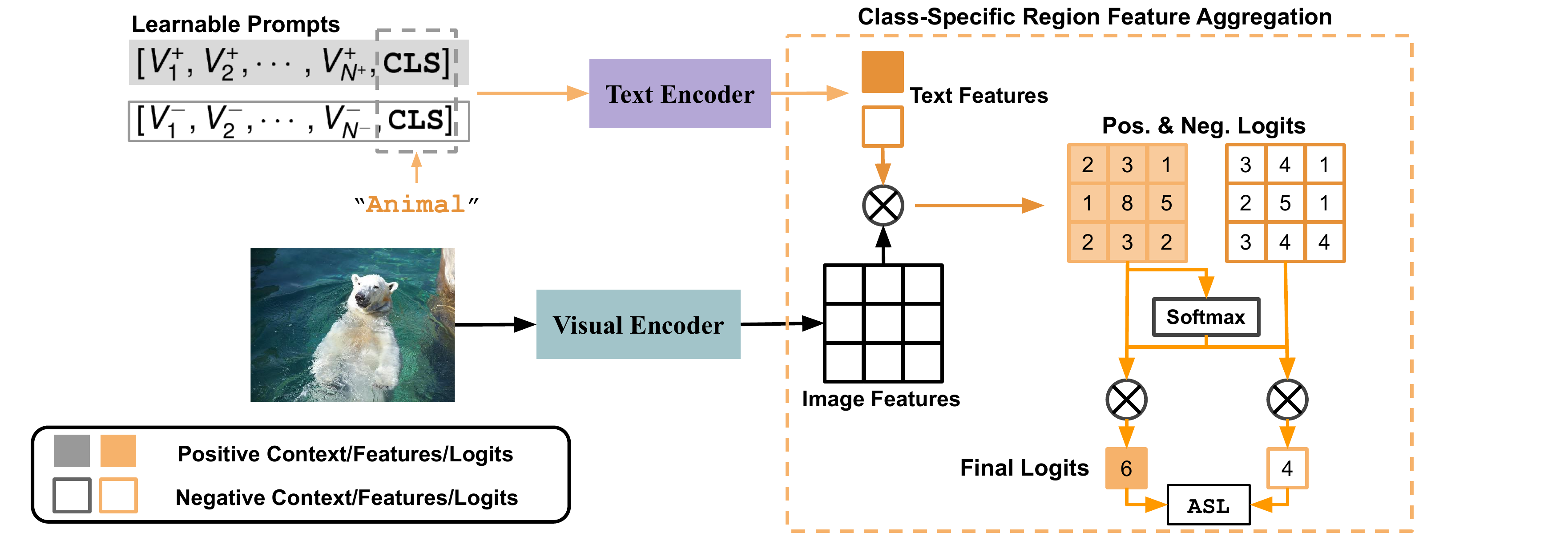}
\end{center} %
   \caption{\small 
   \textbf{Illustration of our proposed approach.} \ours learns a pair of positive and negative prompts to quickly adapt powerful pretrained vision-text encoders to the MLR task. For each class, two prompts generate two contrastive (positive and negative) textual embeddings as the input to the text encoder. Furthermore, we propose \textit{Class-Specific Region Feature Aggregation} to first project each region's feature to the textual space and then aggregate the spatial logits by the magnitude of class-specific semantic responses. During training, we apply the ASL loss~\cite{ridnik2021asymmetric} to optimize learnable prompts while keeping other network components frozen. During inference, we compare the positive and negative logits to make a prediction for each class.
   }\label{fig:model_overview}
   \vspace{-10pt}
\end{figure}

\vspace{1mm}
\textbf{Approach Overview.}
To compensate for insufficient or missing image labels, it is important to learn how the meanings of category names are related to each other, so we can transfer knowledge between related categories. This is usually done by learning an alignment between the visual and textual spaces. However, our dataset is too limited to learn a broad and generalizable mapping. 
We propose \ours to instead leverage the strong alignment of visual and textual feature spaces learned by large-scale vision-language pretraining (CLIP~\cite{radford2019language}) with a light-weight learnable overhead which quickly adapts to the MLR task with limited semantic annotations. Figure~\ref{fig:model_overview} provides an overview of our proposed approach. \ours learns a pair of ``prompt'' contexts in the form of two learnable sequences of word vectors, to provide positive and negative contextual surroundings of a given category name $m$. This generates positive and negative textual features $(F_t^m) _+$  and $(F_t^m)_{-} $ that are fed into the pretrained text encoder. Furthermore, 
to better recognize multiple objects, which can be located at different locations in the image, the spatial aggregation step is modified. We first compute the similarity score of each projected visual feature $F_v^i$ at location $i$ with  $(F_t^m) _+$/$(F_t^m)_{-}$ to obtain prediction logits over regions. For each class,  we perform aggregation of all spatial logits, in which the weight for each logit is determined by its relative magnitude. We call this \textit{Class-Specific Region Feature Aggregation}. During training, we  optimize the learnable prompts via the ASL loss~\cite{ridnik2021asymmetric} while keeping all other network components frozen. During inference,  we directly compare the final positive and negative logits to make a prediction for each label $y_m$.

\vspace{1mm}
\textbf{Dual Learnable Prompts.} Instead of learning a single prompt for a class~\cite{zhou2021learning}, we propose Dual Context Optimization (\ours) which learns two contrastive prompts' contexts for each class. The learnable part in dual prompts carries positive and negative contextual surroundings individually and can be optimized end-to-end from data via binary classification loss. Specifically, we define the pair of prompts given to the text encoder as follows:

\begin{align}
    {\prompt}^{+} &= \big[V_1^{+},V_2^{+}, \cdots ,V_{N^+}^{+}, \texttt{CLS}\big], \\
    {\prompt}^{-} &= \big[V_1^{-},V_2^{-}, \cdots ,V_{N^-}^{-}, \texttt{CLS}\big] 
\end{align}

where each $V$ is a learnable word embedding vector (\textit{e.g.} with dimension 512 in CLIP~\cite{radford2019language}) and $\texttt{CLS}$ is the given category name. $N^{+}$ and $N^{-}$ are the numbers of word tokens learned in the positive  and negative prompts respectively. For simplicity, we set $N^+ = N^-$ in our experiments. We learn a pair of positive and negative prompts for each class (i.e. class-specific prompt pair) when solving MLR with partial labels, and learn a pair of prompts shared for all classes in zero-shot MLR. With a pair of prompts, we compute the binary classification output $p$ with the following form:
\begin{align}
    p = \frac{\exp (<A(E_v(I)), E_t(\prompt^+)>/ \tau)}{\exp (<A(E_v(I)), E_t(\prompt^+)>/ \tau) + \exp (<A(E_v(I)), E_t(\prompt^-)>/ \tau)},
\end{align}
where $<\cdot, \cdot>$ represents cosine similarity and $p$ is the predicted probability for a given (image, label) pair as positive example. $E_v(\cdot)$ and $E_t(\cdot)$ are the visual and textual encoders from the vision-language pretraining. $A(\cdot)$ is our new aggregation function to adaptively reduce the spatial dimension of visual features for each class, which will be discussed next.

\vspace{1mm}
\textbf{Class-Specific Region Feature Aggregation.} In multi-label image recognition, it is common that multiple objects appear in different regions of the image. 
Pooling to produce a single image-level feature vector for all classes gives sub-optimal performance since spatial information is reduced and different objects are mixed.
In this work, we reformulate the last multi-headed attention pooling layer of the visual encoder in CLIP~\cite{radford2019language} and apply class-specific pooling to adaptively aggregate region features in the multi-label setting. The original attention pooling layer in CLIP pools the visual feature map first, and then projects the global feature vector into text space as follows: 
\begin{align}
    \text{AttnPool}& (x)  = \text{Proj}_{v \rightarrow t} (\sum_i \text{softmax}(\frac{q(\bar x) k(x_i)^T}{C}) \cdot v(x_i) ) \nonumber \\
      = \sum_i & \text{softmax}(\frac{q(\bar x) k(x_i)^T}{C}) \cdot \text{Proj}_{v \rightarrow t} (v(x_i)) \nonumber  = \text{Pool}(\text{Proj}_{v \rightarrow t} (v(x_i)) ), 
\end{align}
where $q$, $v$ and $k$ are independent linear embedding layers and $x = E_v(I)$ is the output feature map of the visual encoder. By removing the pooling operation, we can project the visual feature $x_i$  of each region $i$ to the textual space~\cite{zhou2021denseclip}:
\begin{align}
    F_v^i = \text{Proj}_{v \rightarrow t} (v(x_i)).
\end{align}
For each region $i$ and each class $m$, we compute cosine similarity between $F_v^i$ and $(F_t^m) ^{+} = E_t(\prompt^{+})$ as $ S_{i, m}^ + = <F_v^i, (F_t^m)^{+} >$, and compute $S_{i, m}^ -$ in the same way. In order to make a single prediction for the whole image, we aggregate $S_{i,m}^+$ and  $S_{i,m}^-$  into $S_{m}^+$ and  $S_{m}^-$ according to the magnitude of $S_{i,m}^+$, i.e.:
\begin{align}
     S_m^ +  &= A(S_{i,m}^+)  = \sum_i \big(\text{softmax}(S_{i,m}^+) \cdot S_{i,m}^+ \big), \\
    S_m^ - &= A(S_{i,m}^- ) = \sum_i \big( \text{softmax}(S_{i,m}^+) \cdot S_{i,m}^- \big) .
\end{align}
Notably, we do not introduce any new parameters in our re-formulation of the spatial aggregation function. All parameters used to project visual features to the textual space are inherited from the original multi-headed attention pooling layer in CLIP.

\vspace{1mm}
\textbf{Optimization}. 
We apply the Asymmetric Loss (ASL)~\cite{ridnik2021asymmetric} to handle the inherent positive-negative imbalance in the optimization of multi-label recognition. Specially, we compute losses for a positive (image, label) pair  $\mathcal{L}_{+}$ and a negative (image, label) pair  $\mathcal{L}_{-}$ as follows:
\begin{align}
    &\mathcal{L}_{+} = (1-p)^{\gamma_{+}} \log(p), \\
    & \mathcal{L}_{-} = (p_c)^{\gamma_{-}} \log (1 - p_c),
\end{align}
where  $p_c = \max(p-c, 0)$ is the  probability for negative examples shifted by hard thresholding via the margin $c$. We set the hyper-parameters $\gamma_{-} \ge \gamma_{+}$, so that ASL down-weighs
and hard-thresholds easy negative samples. The pair of learnable prompts are updated by back-propagating ASL through the frozen text encoder.

\section{Experiments}\label{sec:experiments}

\subsection{Multi-Label Recognition with Partial Labels}\label{sec:mlc_pl}
\begin{table}
    \begin{center}
     \caption{\small \textbf{Multi-label Recognition on MS-COCO and VOC2007 with partial labels.} \ours achieves the best performance over all SOTA methods. $^*$ indicates previous models using weights pretrained by CLIP~\cite{radford2021learning}}~\label{table:partial_label}
        \resizebox{\linewidth}{!}{
        \begin{tabular}{c| c| c c c c c c c c c | c }
            \Xhline{3\arrayrulewidth} 
            Methods & \cellcolor{yellow!15} \#P & \cellcolor{yellow!15} $10\%$ & $20\%$  & $30\%$ & $40\%$ & $50\%$ & $60\%$ &   $70\%$ &  $80\%$ &  $90\%$  & \cellcolor{yellow!15}Avg. \\
            \Xhline{3\arrayrulewidth} 
            \multicolumn{12}{c}{MS-COCO~\cite{lin2014microsoft}} \\
            \Xhline{3\arrayrulewidth}
            SSGRL~\cite{chen2019learning} & \cellcolor{yellow!15} 64.7M & \cellcolor{yellow!15} 62.5 & 70.5 & 73.2 & 74.5 & 76.3 & 76.5 &  77.1 & 77.9 & 78.4 & \cellcolor{yellow!15} 74.1  \\
            GCN-ML~\cite{chen2019multi} & \cellcolor{yellow!15} 44.9M & \cellcolor{yellow!15} 63.8 & 70.9 & 72.8 & 74.0 & 76.7 & 77.1 & 77.3 & 78.3 & 78.6 & \cellcolor{yellow!15} 74.4 \\
            KGGR~\cite{chen2020knowledge} & \cellcolor{yellow!15} $\ge$ 25M  & \cellcolor{yellow!15} 66.6 & 71.4 & 73.8 & 76.7 & 77.5 & 77.9 & 78.4 & 78.7 & 79.1 & \cellcolor{yellow!15} 75.6 \\
            Curriculum labeling~\cite{durand2019learning} & \cellcolor{yellow!15} $\ge$ 38M & \cellcolor{yellow!15} 26.7 & 31.8 & 51.5 & 65.4 & 70.0 & 71.9 & 74.0 & 77.4 & 78.0 & \cellcolor{yellow!15} 60.7 \\
            Patial BCE~\cite{durand2019learning} & \cellcolor{yellow!15} $\ge$ 38M & \cellcolor{yellow!15} 61.6 & 70.5 & 74.1 &  76.3 & 77.2 &  77.7 & 78.2 &  78.4 & 78.5 &\cellcolor{yellow!15} 74.7\\
            SST~\cite{chen2022structured} & \cellcolor{yellow!15} 33.5M  & \cellcolor{yellow!15} 68.1 &  73.5 & 75.9 & 77.3 & 78.1 & 78.9 & 79.2 & 79.6 & 79.9 & \cellcolor{yellow!15} 76.7 \\
            SST$^*$ &  \cellcolor{yellow!15} 33.5M  & \cellcolor{yellow!15} 69.1 & \underline{78.5} & \underline{79.3} & \underline{79.9} & 80.1 & \underline{80.5} & \underline{81.1} & \underline{80.7} & 80.7 & \cellcolor{yellow!15} 78.9 \\
            SARB~\cite{pu2022semantic} & \cellcolor{yellow!15} 29.6M & \cellcolor{yellow!15} 71.2 & 75.0 & 77.1 & 78.3 & 78.9 & 79.6 & 79.8 &  80.5 & 80.5 & \cellcolor{yellow!15} 77.9 \\
            SARB$^*$ & \cellcolor{yellow!15} 29.6M & \cellcolor{yellow!15} \underline{75.5} & \underline{78.5} & 79.0 & 79.5 & \underline{80.4} & 80.2 & 80.8 & 80.6 & \underline{80.8} & \cellcolor{yellow!15} \underline{79.4} \\
            \ours (ours) & \cellcolor{yellow!15} \textbf{1.3M} & \cellcolor{yellow!15} \textbf{78.7} & \textbf{80.9} & \textbf{81.7} & \textbf{82.0} & \textbf{82.5} & \textbf{82.7} & \textbf{82.8} & \textbf{83.0} & \textbf{83.1} & \cellcolor{yellow!15} \textbf{81.9} \\            
            \Xhline{3\arrayrulewidth} 
              \multicolumn{12}{c}{PASCAL VOC 2007 ~\cite{everingham2010pascal}}\\
            \Xhline{3\arrayrulewidth}
             SSGRL~\cite{chen2019learning} & \cellcolor{yellow!15}66.6M  & \cellcolor{yellow!15} 77.7 & 87.6 & 89.9 & 90.7 & 91.4 & 91.8 & 91.9 & 92.2 & 92.2 & \cellcolor{yellow!15} 89.5\\
            GCN-ML~\cite{chen2019multi} & \cellcolor{yellow!15}44.9M & \cellcolor{yellow!15} 74.5 & 87.4 & 89.7 & 90.7 & 91.0 & 91.3 & 91.5 & 91.8 & 92.0 & \cellcolor{yellow!15} 88.9\\
            KGGR~\cite{chen2020knowledge} & \cellcolor{yellow!15}$\ge$ 25M  & \cellcolor{yellow!15} 81.3 & 88.1 & 89.9 & 90.4 & 91.2 & 91.3 & 91.5 & 91.6 & 91.8 & \cellcolor{yellow!15} 89.7 \\
            Curriculum labeling~\cite{durand2019learning} & \cellcolor{yellow!15} $\ge$ 38M  & \cellcolor{yellow!15} 44.7 & 76.8 & 88.6 & 90.2 & 90.7 & 91.1 & 91.6 & 91.7 & 91.9 & \cellcolor{yellow!15} 84.1\\
            Patial BCE~\cite{durand2019learning} & \cellcolor{yellow!15} $\ge$ 38M & \cellcolor{yellow!15} 80.7 & 88.4 & 89.9 &  90.7 & 91.2 & 91.8 & 92.3 & 92.4 & 92.5  & \cellcolor{yellow!15} 90.0 \\
            SST~\cite{chen2022structured} & \cellcolor{yellow!15} 32.4M & \cellcolor{yellow!15} 81.5 & 89.0 & 90.3 & 91.0 & 91.6 &  92.0 & 92.5 & 92.6 & 92.7 & \cellcolor{yellow!15} 90.4 \\
            SARB~\cite{pu2022semantic} & \cellcolor{yellow!15} 29.6M & \cellcolor{yellow!15} \underline{83.5} & \underline{88.6} & \underline{90.7} & \underline{91.4} & \underline{91.9} & \underline{92.2} & \underline{92.6} & \underline{92.8} & \underline{92.9} & \cellcolor{yellow!15} \underline{90.7} \\
            \ours (ours) &\cellcolor{yellow!15} \textbf{0.3M} & \cellcolor{yellow!15} \textbf{90.3} & \textbf{92.2} & \textbf{92.8} & \textbf{93.3} & \textbf{93.6} & \textbf{93.9} & \textbf{94.0} & \textbf{94.1} & \textbf{94.2} & \cellcolor{yellow!15} \textbf{93.2} \\ 
             \Xhline{3\arrayrulewidth} 
        \end{tabular}
        } 
    \end{center}
\vspace{-10pt}
\end{table}

\textbf{Datasets.} 
We conduct experiments on MS-COCO~\cite{lin2014microsoft} and VOC2007~\cite{everingham2010pascal} to evaluate multi-label recognition with partial labels. MS-COCO~\cite{lin2014microsoft} contains 80 common object categories and we use the official \texttt{train2014} (82K images) and \texttt{val2014} (40K images) splits for training and test. VOC2007~\cite{everingham2010pascal} contains 20 object categories and we use the official \texttt{trainval} (5K images) and \texttt{test} (5K images) splits for training and test. To create the training set with partial labels, we randomly mask out labels from the fully annotated training set\footnote{The difference in performance is within $1.0\%$ of independent runs.} and use the remaining labels for training by following standard practice~\cite{chen2022structured,durand2019learning,pu2022semantic}. In this work, we vary the proportion of kept labels from $10\%$ to $90\%$~\cite{chen2022structured,pu2022semantic}.

\textbf{Evaluation.} 
On MS-COCO and VOC2007 datasets, we follow \cite{chen2022structured,durand2019learning, pu2022semantic} to report the mean average precision (mAP) for each proportion of labels available for optimization (from $10\%$ to $90\%$) and its average value for all proportions.
We count the learnable parameters (\#P) of each baseline and \ours to measure the complexity of optimization\footnote{For baselines without public released implementation, we only measure the major part of the learnable parameters based on description in their papers. (indicated as \#P $\ge$ [a value] in Table~\ref{table:partial_label}-\ref{table:zsl_nus_wide})}. 
We also report the per-class and the average overall precision (CP and OP), recall (CR and OR), and F1 (CF1 and OF1) of \ours under different proportions of labels for training in the supplementary material due to the page limit.

\textbf{Implementation.} We adopt ResNet-101~\cite{he2016deep} as the visual encoder in all baselines and \ours for input resolution 448$\times$448,  and use the same Transformer~\cite{radford2019language,vaswani2017attention} in CLIP~\cite{radford2021learning} as the text encoder. The visual and text encoders are initialized from the CLIP pretrained model and kept frozen during optimization. For each class/label, we learn two independent context vectors with 16 context tokens (N = 16) following \cite{zhou2021learning}, which is the only learnable part in \ours. We use the SGD optimizer with an initial rate of 0.002 which is decayed by the cosine annealing rule. We train context vectors for 50 epochs with a batch-size 32/8 for MS-COCO/VOC2007, respectively. For ASL loss, we choose $\gamma_{+} = 1$, $\gamma_{-}=2$ and $c=0.05$ via validation. The training is done with one RTX A6000.

\textbf{Baselines.} To evaluate the effectiveness of \ours, we compare with the following baselines: (1).SSGRL~\cite{chen2019learning}, GCN-ML~\cite{chen2019multi} and KGGR~\cite{chen2020knowledge} adopt graph neural networks to model label dependencies. We follow \cite{chen2022structured} to report their performance in the partial-label setting. (2). Curriculum labeling~\cite{durand2019learning} and  SST~\cite{chen2022structured} generate pseudo labels for unknown labels. (3). Partial BCE~\cite{durand2019learning} uses a normalized BCE loss to better exploit partial labels. (4).SARB~\cite{pu2022semantic} blends category-specific representation across different images to transfer information of known labels to complement unknown labels.

\textbf{Results.} 
Table~\ref{table:partial_label} shows the comparison of mAP between \ours and all baselines optimized with $10\%$ to $90\%$ of labels. For the two most recent works (SST~\cite{chen2022structured} and SARB~\cite{pu2022semantic}), we further substitute the ImageNet pretrained weights~\cite{he2016deep} with the CLIP pretrained weights~\cite{radford2021learning} when initializing of their visual encoders, which results in $\text{SST}^*$ and $\text{SARB}^*$ in Table~\ref{table:partial_label}. Since we learn class-specific prompts, \ours on MS-COCO adopts more learnable parameters than VOC2007. Our proposed \ours achieves the best performance across all  proportions of labels available during the training with the smallest learnable overhead (1.3M vs. 29.6M in $\text{SARB}^*$ on MS-COCO and 0.3M vs. 29.6M in $\text{SARB}^*$ on VOC2007). Notably, \ours yields a great improvement over the second-best method, $3.2\%$ on MS-COCO and $6.8\%$ on VOC2007,  especially when only providing $10\%$ of labels during the training. This indicates that \ours can quickly adapt to the multi-label recognition task with a few labels by taking advantage of the powerful vision-language pretraining.

\subsection{Zero-shot Multi-Label Recognition}
\textbf{Datasets.} 
Following \cite{ben2021semantic, huynh2020shared}, we conduct experiments on MS-COCO~\cite{lin2014microsoft} and NUS-WIDE~\cite{chua2009nus} to perform zero-shot multi-label recognition. On MS-COCO, we follow \cite{bansal2018zero, ben2021semantic} to split the dataset into 48 seen classes and 17 unseen classes. NUS-WIDE~\cite{chua2009nus} dataset includes 270K images. Following \cite{ben2021semantic, huynh2020shared} we use 81 human-annotated categories as unseen classes and an additional set of 925 labels obtained from Flickr tags as seen classes. 

\textbf{Evaluation.}
We follow \cite{ben2021semantic} and report precision, recall, and F1 score at Top-3 predictions in each image on MS-COCO. We also follow \cite{ben2021semantic, huynh2020shared} to report mAP over all categories as well as precision, recall, and F1 score at Top-3 and Top-5 predictions in each image on NUS-WIDE. We evaluate all methods with both zero-shot setting (test only on unseen classes) and generalized zero-shot setting (test on both seen and unseen classes).

\textbf{Implementation.} We adopt ResNet-50~\cite{he2016deep} similar to \cite{ben2021semantic} as the visual encoder in \ours for input resolution 224. Instead of learning class-specific prompts, we learn the class-agnostic context vectors with 64 context tokens (N = 64) for all classes, which is the only learnable part in \ours. We optimize context vectors for 50 epochs with a batch-size 32/192 for MS-COCO/NUS-WIDE, respectively. Other implementation details are the same with Sec.~\ref{sec:mlc_pl}

\textbf{Baselines.}
To evaluate the effectiveness of \ours in the zero-shot setting, we compare with the following baselines: (1). CONSE~\cite{norouzi2013zero} adopts an ensemble of classifiers for unseen classes. (2). LabelEM~\cite{akata2015label} learns a joint image-label embedding. (3). Fast0Tag~\cite{zhang2016fast} and SDL~\cite{ben2021semantic} estimate one or multiple diverse principal directions of the input images. (4). Deep0Tag~\cite{rahman2018deep} and LESA~\cite{huynh2020shared} estimate the relevant regions via region proposals and attention techniques respectively. (5). BiAM~\cite{narayan2021discriminative} enhances the region-based features to minimize inter-class feature entanglement.

\begin{table}
    \begin{center}
     \caption{\small{\textbf{Zero-Shot Multi-Label Recognition on MS-COCO\cite{lin2014microsoft}}. \ours achieves the best F1 score in both ZSL and GZSL settings.}}~\label{tab:zero_shot_mscoco}
    
        \resizebox{0.68\linewidth}{!}{
        \begin{tabular}{c c c c c c c c}
        \Xhline{3\arrayrulewidth} 
        \multirow{2}{*}{Methods} & \cellcolor{yellow!15}  & \multicolumn{3}{c}{ZSL} & \multicolumn{3}{c}{GZSL} \\
        &  \cellcolor{yellow!15}\multirow{-2}{*}{\#P}  & \textbf{P} & \textbf{R} & \cellcolor{yellow!15} \textbf{F1} & \textbf{P} & \textbf{R} &\cellcolor{yellow!15} \textbf{F1} \\ 
           \Xhline{3\arrayrulewidth} 
           CONSE~\cite{norouzi2013zero} &  \cellcolor{yellow!15}  -   & 11.4 & 28.3 &  \cellcolor{yellow!15}  16.2  & 23.8 & 28.8 &  \cellcolor{yellow!15} 26.1 \\
           Fast0Tag~\cite{zhang2016fast} &  \cellcolor{yellow!15}  0.61M    & 24.7 & 61.4 &  \cellcolor{yellow!15}  25.3 & 38.5 & 46.5 &  \cellcolor{yellow!15}  42.1 \\
           Deep0Tag~\cite{rahman2018deep}  &  \cellcolor{yellow!15}  $\ge$ 23M   & \underline{26.5} & \underline{65.9} &  \cellcolor{yellow!15} \underline{37.8} &  43.2 &  52.2 &  \cellcolor{yellow!15}  47.3 \\
           SDL (M=2)~\cite{ben2021semantic}  &  \cellcolor{yellow!15}  30.6M   &  26.3 & 65.3 &  \cellcolor{yellow!15}  37.5 &  \textbf{59.0} & \underline{60.8} &  \cellcolor{yellow!15}  \underline{59.9} \\
           \ours (ours) &   \cellcolor{yellow!15}  \textbf{0.02M} & \textbf{35.3} & \textbf{87.6} & \cellcolor{yellow!15} \textbf{50.3}  &	\underline{58.4} & \textbf{68.1} & \cellcolor{yellow!15}  \textbf{62.9}	 \\
            \Xhline{3\arrayrulewidth} 
        \end{tabular}
        } 
    \end{center}
\vspace{-20pt}
\end{table}

\begin{table}
    \begin{center}
     \caption{\small \textbf{Zero-Shot Multi-label Recognition on NUS-WIDE~\cite{chua2009nus}.} \ours achieves the best F1 score over all SOTA methods at Top-3/Top-5 predictions in both ZSL and GZSL settings.  }~\label{table:zsl_nus_wide}
        \resizebox{0.83\linewidth}{!}{
        \begin{tabular}{c c c c c c c c c}
        \Xhline{3\arrayrulewidth} 
        \multirow{2}{*}{Methods} & \cellcolor{yellow!15} & \multicolumn{3}{c}{Top-3} & \multicolumn{3}{c}{Top-5} & \cellcolor{yellow!15}  \\
        & \cellcolor{yellow!15} \multirow{-2}{*}{\#P} & \textbf{P} & \textbf{R} & \cellcolor{yellow!15} \textbf{F1} & \textbf{P} & \textbf{R} & \cellcolor{yellow!15}\textbf{F1} & \cellcolor{yellow!15} \multirow{-2}{*}{mAP}\\
        \Xhline{3\arrayrulewidth} 
         \multicolumn{9}{c}{Zero-Shot Learning (ZSL)}  \\
 \Xhline{3\arrayrulewidth} 
CONSE~\cite{norouzi2013zero} & \cellcolor{yellow!15} - & 17.5 & 28.0 &  \cellcolor{yellow!15} 21.6 & 13.9 & 37.0 & \cellcolor{yellow!15} 20.2 &\cellcolor{yellow!15} 9.4 \\ 
LabelEM~\cite{akata2015label} & \cellcolor{yellow!15} - & 15.6 & 25.0 & \cellcolor{yellow!15} 19.2 & 13.4 & 35.7 & \cellcolor{yellow!15} 19.5 & \cellcolor{yellow!15} 7.1 \\ 
Fast0Tag~\cite{zhang2016fast} & \cellcolor{yellow!15} 0.61M  &   22.6 &  36.2 & \cellcolor{yellow!15} 27.8 &  18.2 & 48.4 & \cellcolor{yellow!15} 26.4 & \cellcolor{yellow!15} 15.1 \\
One Attention per Label~\cite{Kim2018}  & $\cellcolor{yellow!15} \ge$ 12.8M &  20.9 & 33.5 & \cellcolor{yellow!15} 25.8 & 16.2 & 43.2 & \cellcolor{yellow!15} 23.6 & \cellcolor{yellow!15} 10.4 \\
LESA (M=10)~\cite{huynh2020shared} & \cellcolor{yellow!15} $\ge$ 0.45M &  \underline{25.7} &  41.1 & \cellcolor{yellow!15} 31.6 & \underline{19.7} & 52.5 & \cellcolor{yellow!15} 28.7 & \cellcolor{yellow!15} 19.4  \\
BiAM~\cite{narayan2021discriminative} & \cellcolor{yellow!15} 3.8M & -- & -- & \cellcolor{yellow!15} \underline{33.1} & -- & -- & \cellcolor{yellow!15} \underline{30.7} & \cellcolor{yellow!15} \underline{26.3} \\
SDL (M=7)~\cite{ben2021semantic} & \cellcolor{yellow!15} 33.6M  & 24.2 & \underline{41.3} & \cellcolor{yellow!15} 30.5 & 18.8 & \underline{53.4} & \cellcolor{yellow!15} 27.8 &\cellcolor{yellow!15} 25.9  \\
\ours (ours) & \cellcolor{yellow!15} \textbf{0.02M}  &  \textbf{37.3} &	\textbf{46.2} & \cellcolor{yellow!15} \textbf{41.3} & \textbf{28.7} & \textbf{59.3} & \cellcolor{yellow!15} \textbf{38.7} & \cellcolor{yellow!15} \textbf{43.6}  \\
             \Xhline{3\arrayrulewidth} 
             \multicolumn{9}{c}{Generalized Zero-Shot Learning (GZSL)}  \\
             \Xhline{3\arrayrulewidth} 
CONSE~\cite{norouzi2013zero} & \cellcolor{yellow!15} -  & 11.5 & 5.1 & \cellcolor{yellow!15}7.0 & 9.6 & 7.1 &  \cellcolor{yellow!15} 8.1 &  \cellcolor{yellow!15}  2.1  \\
LabelEM~\cite{akata2015label} & \cellcolor{yellow!15} - & 15.5 & 6.8 & \cellcolor{yellow!15}9.5 & 13.4 & 9.8 & \cellcolor{yellow!15}  11.3 &\cellcolor{yellow!15}  2.2  \\
Fast0Tag~\cite{zhang2016fast} & \cellcolor{yellow!15} 0.61M & 18.8 & 8.3 & \cellcolor{yellow!15}11.5 & 15.9 & 11.7 &  \cellcolor{yellow!15}  13.5 & \cellcolor{yellow!15}  3.7  \\
One Attention per Label~\cite{Kim2018}  & \cellcolor{yellow!15} $\ge$ 12.8M & 17.9 & 7.9 & \cellcolor{yellow!15}10.9 & 15.6 & 11.5 & \cellcolor{yellow!15}  13.2 & \cellcolor{yellow!15}  3.7 \\
LESA (M=10)~\cite{huynh2020shared} & $\cellcolor{yellow!15}\ge$ 0.45M & 23.6 & 10.4 & \cellcolor{yellow!15}14.4 & 19.8 & 14.6 & \cellcolor{yellow!15} 16.8 &\cellcolor{yellow!15}   5.6 \\
BiAM~\cite{narayan2021discriminative} & \cellcolor{yellow!15} 3.8M & -- & -- & \cellcolor{yellow!15} 16.1 & -- & -- & \cellcolor{yellow!15} 19.0 & \cellcolor{yellow!15} 9.3 \\
SDL (M=7)~\cite{ben2021semantic} & \cellcolor{yellow!15}33.6M   & \underline{27.7} & \textbf{13.9} &\cellcolor{yellow!15} \underline{18.5} & \underline{23.0} & \textbf{19.3} & \cellcolor{yellow!15}  \underline{21.0} & \cellcolor{yellow!15} \textbf{ 12.1} \\
\ours (ours) & \cellcolor{yellow!15}\textbf{ 0.02M  } & \textbf{31.9} & \textbf{13.9} & \cellcolor{yellow!15} \textbf{19.4} & \textbf{26.2} & 19.1 & \cellcolor{yellow!15} \textbf{ 22.1} & \cellcolor{yellow!15} \underline{12.0} \\
           \Xhline{3\arrayrulewidth} 
        \end{tabular}
        } 
    \end{center}
\vspace{-15pt}
\end{table}

\textbf{Results.} 
Table~\ref{tab:zero_shot_mscoco}-\ref{table:zsl_nus_wide} shows the comparison between \ours and all SOTA methods of zero-shot learning and generalized zero-shot learning on MS-COCO and NUS-WIDE datasets. \ours achieves the best F1 score in all cases with a very light learnable overhead (0.02M) and improves the performance of zero-shot learning (unseen labels) with a significant margin: F1 score improves by 12.5 @Top-3 on MS-COCO, and by 10.8 @Top-3 and 10.9 @Top-5 on NUS-WIDE. This shows the power of exploiting the pretrained alignment of textual and visual spaces in CLIP via \ours to solve multi-label recognition.

\subsection{Ablation Studies}

\begin{table}
    \begin{center}
     \caption{\small \textbf{Comparison among methods on MS-COCO using partial labels with the same initialization. All methods use parameters pretrained by CLIP~\cite{radford2021learning}.}}~\label{table:semantic_guide}
        \resizebox{0.65\linewidth}{!}{
        \begin{tabular}{c | c |c c c c c  }
            \Xhline{3\arrayrulewidth} 
            Method &  Text Supervision & $10\%$  & $30\%$ & $50\%$ & $70\%$ & $90\%$ \\
            \Xhline{3\arrayrulewidth}  
            Discrete Label & \xmark & 70.6 & 75.1 &  76.5 & 77.3 & 78.0 \\
            SST & \cmark &  69.1 & 79.3 & 80.1 & 81.1 & 80.7\\
            SARB & \cmark &  75.5 & 79.0 & 80.4 & 80.8 & 80.8\\
            \ours & \cmark & \textbf{ 78.4} &\textbf{ 81.0} & \textbf{82.0} & \textbf{82.5}  & \textbf{82.8 } \\
             \Xhline{3\arrayrulewidth} 
        \end{tabular}
        } 
    \end{center}
\vspace{-10pt}
\end{table}
\textbf{Effectiveness of Text Supervision.} To show the effectiveness of text supervision from label space, we compare the model learned with discrete label space (``Discrete Label'') with three methods (SST~\cite{chen2022structured}, SARB~\cite{pu2022semantic}] and  \ours) which introduce the textual space to utilize the contextual correlation of labels in Table~\ref{table:semantic_guide}. We find that methods with text supervision usually perform better than the method only using discrete labels. However, when the semantic annotations are limited, text supervision sometimes yields worse performance (e.g. mAP of SST is $1.5\%$ lower than Discrete Labels with only $10\%$ of labels). By adopting the well-pretrained visual-textual alignment, \ours achieves a great performance (\textit{e.g.} $7.8\%$ higher than Discrete Labels with $10\%$ of labels) and quickly adapts to the dataset even with limited labels.

\begin{table}
    \begin{center}
    \vspace{-10pt}
     \caption{\small \textbf{Ablation on Linguistic Inputs for Zero-Shot Learning of MS-COCO.}}~\label{table:ablation_prompt}
        \resizebox{\linewidth}{!}{
        \begin{tabular}{c c c c c c  c c c  }
            \Xhline{3\arrayrulewidth} 
            & \multirow{2}{*}{Linguistic Input} & \multirow{2}{*}{\#P}  & \multicolumn{3}{c}{ZSL} & \multicolumn{3}{c}{GZSL}  \\
            & & & \textbf{P} & \textbf{R} & \textbf{F1} & \textbf{P} & \textbf{R} & \textbf{F1} \\
            \Xhline{3\arrayrulewidth} 
            \scriptsize{M0} &Contextless Classname & 0 & 5.2 & 12.9 & 7.4 & 3.5 & 4.1 & 3.8 \\
            \scriptsize{M1} &Hand-crafted Pos./Neg. Templates + Classname & 0 & 25.6 & 63.6 & 36.5 &  31.0 & 36.2 & 33.4  \\
            \scriptsize{M2} &Pos. Learnable Prompt + Classname ($N$=64) & 0.01M & 31.2 & 77.5 & 44.5 & 55.7 & 65.0 & 60.0 \\
            \scriptsize{M3} &Neg. Learnable Prompt + Classname ($N$=64) & 0.01M & 9.3 & 23.0 & 13.2 & 2.6 & 3.0 & 2.8 \\
            \hline
            \scriptsize{M4} &Dual Learnable Prompts + Classname ($N$=64) & 0.02M & 35.3 & 87.6 & 50.3  &	\textbf{58.4} & \textbf{68.1} & \textbf{62.9}  \\
            \scriptsize{M5} &Dual Learnable Prompts + Classname ($N$=32)  & 0.01M & \textbf{35.8} & \textbf{88.9} &	\textbf{51.0} & 57.4 &  67.0 &  61.9  \\
             \Xhline{3\arrayrulewidth} 
        \end{tabular}
        } 
    \end{center}
\end{table}

\textbf{Ablation of Prompt Design.}  We compare our proposed dual learnable prompts with two hand-crafted prompts and one prompt learning method on the MS-COCO dataset with the zero-shot setting (see Table~\ref{table:ablation_prompt}). Hand-crafted prompts can use either contextless class names~\cite{li2017learning} or manually designed prompt templates. In our experiments, we carefully choose the positive and negative prompt templates as ``a photo of a [classname]'' and ``a photo without a [classname]''. In contrast with performing the binary classification for each class with dual learnable prompts as the input, we also experiment with learning a single prompt of positive or negative contexts and use a chosen threshold (0.5 in our experiment) to make the prediction for each class. 
As we can see, the single positive prompt learning method (M2) performs better than non-learnable methods (M0 and M1), and a single negative learnable prompt (M3) achieves much worse accuracy than its positive counterpart (M2).  
However, when we include both positive and negative prompts, dual prompts (M4) performs even better than a single prompt, which indicates that \ours learns complementary and beneficial information in the dual prompt pair. 
To keep the same amount of learnable parameters as in single prompt settings, we also halve the token size (M5), and find that \ours still outperforms two single prompts in M2 and M3 by large gaps, demonstrating the effectiveness of our dual-prompt design.

\begin{table}
    \begin{center}
     \caption{\small \textbf{Comparison between multi-headed attention and class-specific feature aggregation on MS-COCO} }~\label{table:mha_vs_conv_proj}
        \resizebox{0.9\linewidth}{!}{
        \begin{tabular}{c| c c c | c c c c c  }
            \Xhline{3\arrayrulewidth} 
            Visual Aggregation & Finetune. & Train Res. & Test Res. & $10\%$  & $30\%$ & $50\%$ & $70\%$ & $90\%$  \\
            \Xhline{3\arrayrulewidth} 
              \multirow{4}{*}{\makecell{Multi-Headed \\ Attention}}&  \xmark & 224 & 224 & 70.4 & 74.1 & 74.8 & 75.4 & 75.7 \\
            &  \xmark  & 224 & 448 & 65.9 & 70.2 & 71.2 & 72.0 & 72.1 \\
            &  \xmark  & 448 & 448 & 72.1 & 75.5 & 76.5 & 77.1 & 77.3 \\
            &  \cmark  & 448 & 448 & 74.1 & 77.6 & 78.2 & 78.5 & 78.4 \\
            \hline
              \multirow{3}{*}{\makecell{Class-Specific \\ Feature Aggregation \\ (\ours)}} & \xmark  & 224 & 224 & 73.1 & 76.4 & 77.7 & 78.2 & 78.4 \\
            & \xmark  & 224 & 448 & 76.0 & 78.1 & 79.5 & 80.3 & 80.5 \\
            & \xmark  & 448 & 448 &  \textbf{78.4} & \textbf{81.1} & \textbf{82.0} & \textbf{82.5} & \textbf{82.8} \\
             \Xhline{3\arrayrulewidth} 
        \end{tabular}
        } 
    \end{center}
\vspace{-10pt}
\end{table}
\textbf{Multi-Headed Attention vs. Class-Specific Region Aggregation.} 
In Table~\ref{table:mha_vs_conv_proj}, we compare the adaptive ability of these two visual aggregation methods when training/testing with a larger resolution (see Table~\ref{table:mha_vs_conv_proj}), which is crucial in multi-label recognition as spatial details matter. For a fair comparison, we only replace the class-specific region aggregation in \ours with the original multi-headed attention layer in CLIP~\cite{radford2019language} at the end of the visual encoder. We adaptively resize the input feature map to match the input dimension of the multi-headed attention layer.  

\begin{wrapfigure}{r}{0.4\textwidth}
\centering
\vspace{-10pt}
\includegraphics[width=0.99\linewidth]{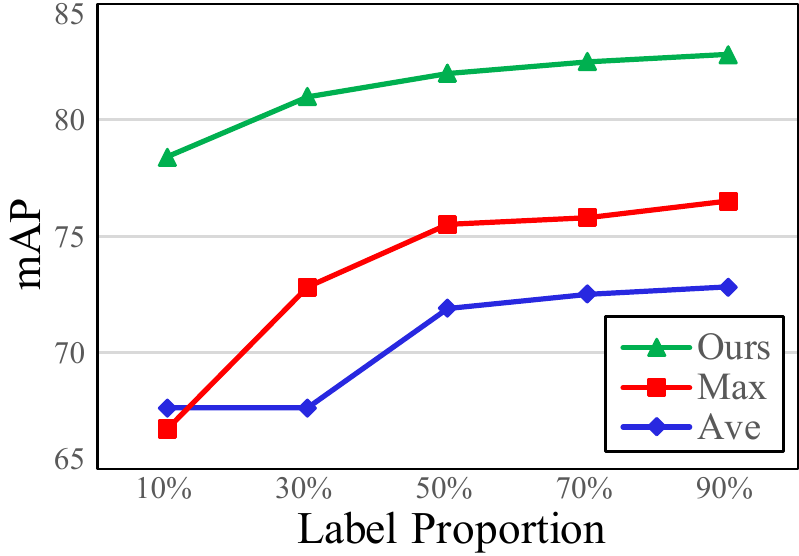}
    \caption{\small{Comparison among different aggregations on MS-COCO using partial labels.}}
\label{fig:aggre}
\vspace{-0.3cm}
\end{wrapfigure}

As shown in Table~\ref{table:mha_vs_conv_proj}, multi-headed attention is bonded to the pre-training image resolution (224 in CLIP), while our class-specific region aggregation benefits from the increased input resolution either during training or in inference. Our class-specific feature aggregation uses original weights, but actually performs better than finetuning the original multi-headed attention layer.

\textbf{Ablation of Aggregation Function.} We experiment with different functions to aggregate the regional logits for each class in Fig.~\ref{fig:aggre}. We compute final logits in three ways: (1) taking the average of logits at all spatial locations (``Ave''), (2) taking the region with the largest positive logit (``Max''), and (3) generating aggregating weights for all spatial locations via a softmax function over the positive logits (``Ours'').  ``Max'' performs better than ``Ave'', which indicates the regional feature is more informative than the global feature in multi-label recognition. Furthermore, by taking account of both the regional and the global features, ``Ours'' gives the best performance.

\vspace{-10pt}
\section{Conclusion}
In this paper,  we propose a unified framework, \ours, for two types of multi-label recognition with limited annotations. It utilizes the powerful vision-language pretraining from a large-scale dataset. 
By introducing a lightweight learnable overhead, it can quickly adapt to solve multi-label recognition after receiving a small amount of labels. 
In \ours, we learn a pair of positive and negative prompts  followed by the target class name as the linguistic input. Furthermore, to better aggregate visual region features for each class, we reformulate the original visual attention in the pretraining model as a class-specific region feature aggregation. We conduct extensive experiments for both partial-label MLR and Zero-Shot MLR across MS-COCO, VOC2007, and NUS-WIDE datasets showing the efficacy of our proposed approach over state-of-the-art methods.

\textbf{Limitations.}
Since the vision-language pretraining adopts a large Transformer-based language model and all labels need to be feed-forward through the text encoder, the large language model limits the size of the label set. Also, compared to training the model with both seen and unseen labels, we still get worse performance for the zero-shot unseen classes even though we have used  400M auxiliary samples in the pretraining. This highlights the difficulty of zero-shot MLR.

\textbf{Negative Societal Impacts.}  Negative impacts of our research are difficult to predict, however, it shares many of the pitfalls associated with deep learning models. These include susceptibility to adversarial attacks and data poisoning, dataset bias, and lack of interpretability. Other risks associated with the deployment of computer vision systems include privacy violations when images are captured without consent, or used to track individuals for profit, or increased automation resulting in job losses. While we believe that these issues should be mitigated, they are beyond the scope of this paper. Furthermore, we should be cautious of the result of failures of the system which could impact the performance/user experience of the high-level AI systems based on our research. 

\newpage

\small
{
}

\appendix

\section{Different Prompt Length}

\begin{wrapfigure}{r}{0.5\textwidth}
\centering
\vspace{-10pt}
 \includegraphics[width=1\linewidth]{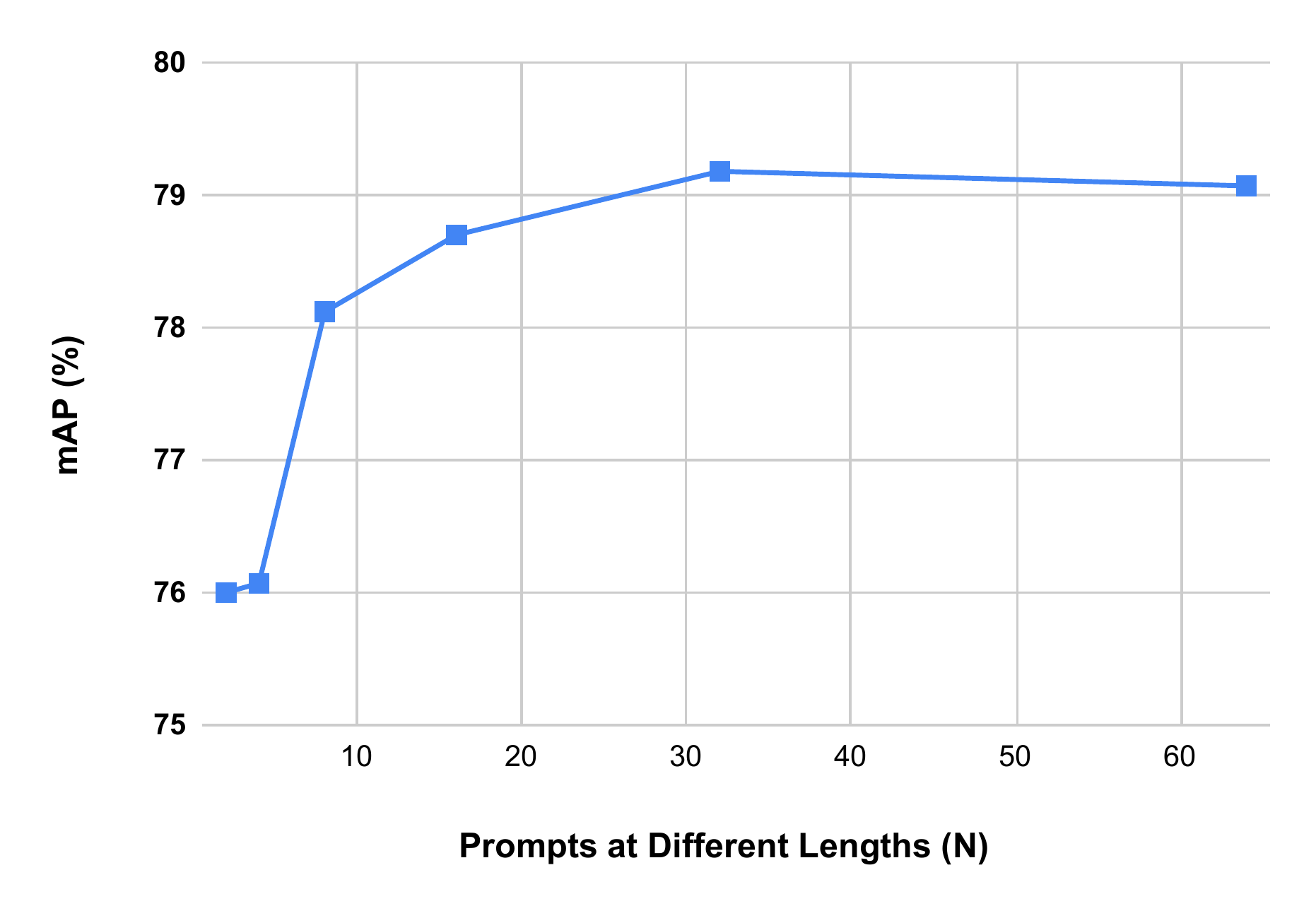}     \caption{\small \textbf{MLR with Partial Labels at Different Prompt Length on MS-COCO~\cite{lin2014microsoft}} }
\label{fig:pl_mlr_N}
\vspace{-50pt}
\end{wrapfigure}

\begin{figure}[h]
\vspace{40pt}
    \centering
    \includegraphics[width=0.48\linewidth]{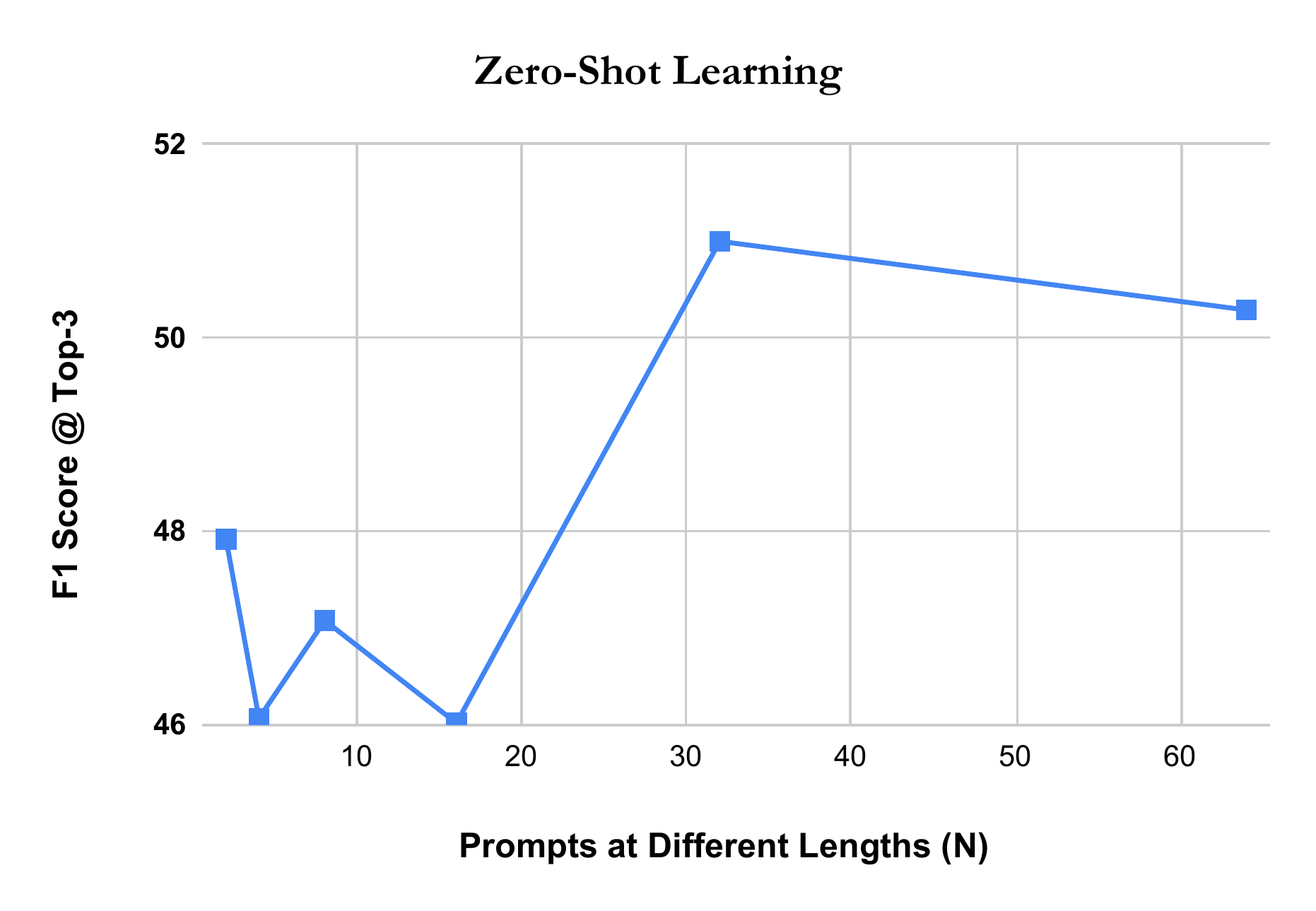} 
    \hfill
  \includegraphics[width=0.51\linewidth]{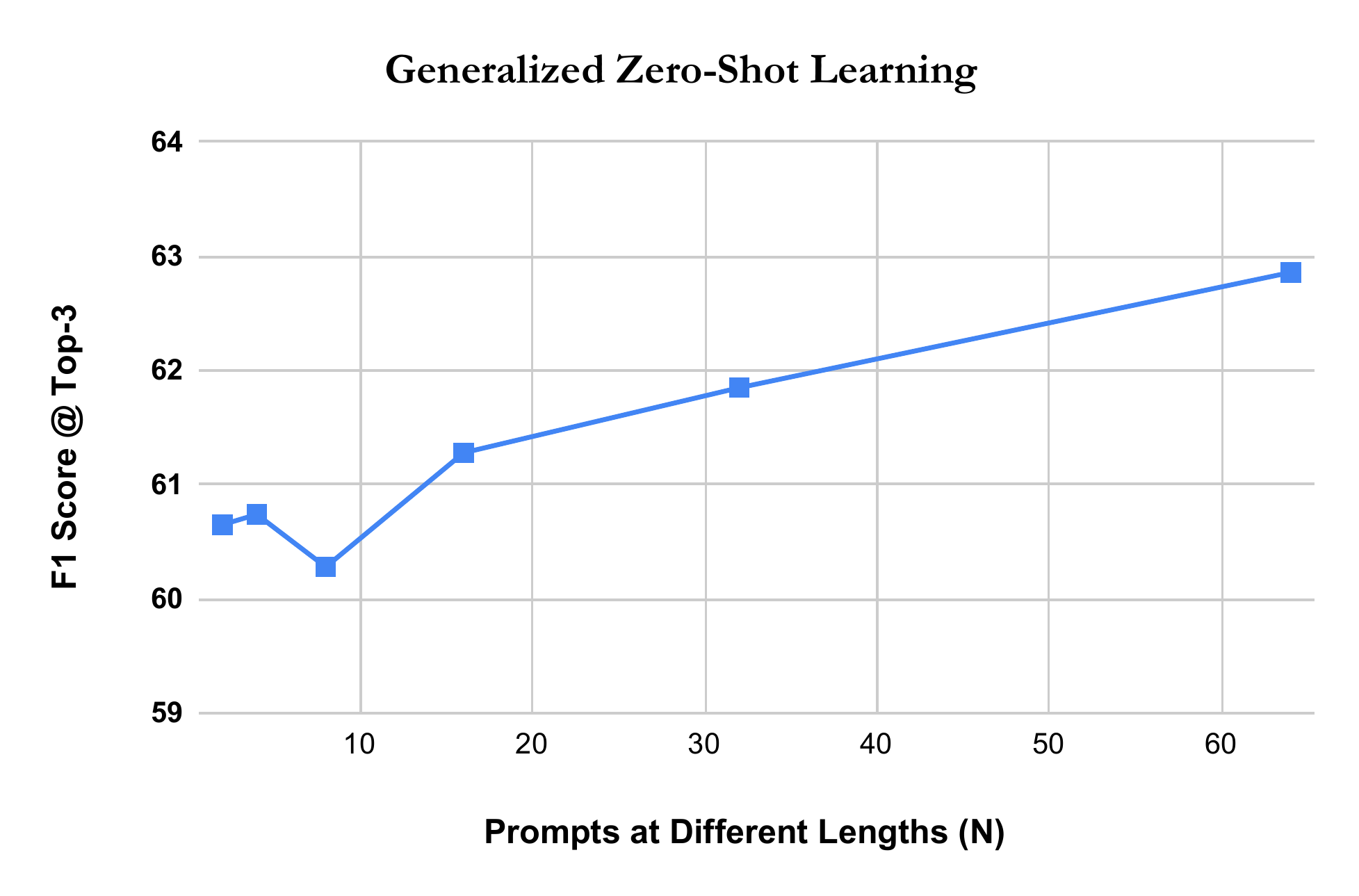} 

    \caption{\small\textbf{Zero-Shot MLR with Different Prompt Length on MS-COCO~\cite{lin2014microsoft}} }
    \label{fig:zs_mlr_N} 
\end{figure}

We have provided the comparison of the performance of \ours with different lengths of prompt context (i.e. $N=2, 4, 6, 8, 16, 32, 64$) in all three different experiment scenarios (see Fig.~\ref{fig:pl_mlr_N} and \ref{fig:zs_mlr_N}). In MLR with partial labels, we learn class-specific prompts and thus \ours performs good when $N$ is small, such as 8, 16. For zero-shot learning in MLR, we learn uniform prompts shared by all classes and it requires larger $N$ (e.g. 32 or 64) for good performance. In the main paper, we use $N=16$ for all experiments of MLR with partial labels and use $N=32$ for experiments in zero-shot learning.

\section{Performance on the Full Dataset}
We also finetune the visual backbone on the full multi-labeled recognition dataset MS-COCO and achieve mAP $85.8\%$ with ResNet 101 and input resolution 448, comparing to $85.0\%$ achieved by the same setting using ASL~\cite{ridnik2021asymmetric}.

\section{Full performance of MLR with Partial Labels}
In this section, we provide the average per-class and average overall precisions (CP and OP), recalls (CR and oR) and F1 scores (CF1 and OF1) of \ours in the experiment of MLR with Partial Labels on MS-COCO~\cite{lin2014microsoft} and VOC2007~\cite{everingham2010pascal} (see Table~\ref{table:ms_coco_full} and \ref{table:voc_full} in supplementary material) as a supplementary for Table 1 in the main paper. 

\begin{table}
    \begin{center}
     \caption{\small \textbf{Performance of MLR with partial labels on MS-COCO}}~\label{table:ms_coco_full}
        \resizebox{0.9\linewidth}{!}{
        \begin{tabular}{c c c c c c c c }
            \Xhline{3\arrayrulewidth} 
            Amount of Labels & CP & CR & CF1& OP & OR & OF1 & mAP \\
            \Xhline{3\arrayrulewidth} 
            10\% &  69.1 &	77.5 &	72.6 &	71.4 &	81.6 & 	76.2 &	78.7 \\
            20\% & 70.1 & 79.4 & 74.2 &	72.1 & 83.0 & 77.2 & 80.9 \\
            30\% & 71.2. & 	80.1 &	75.1. &	72.9. &	83.5 &	77.8 &	81.7 \\
            40\% & 71.3 & 80.2 & 75.2 & 73.2  & 83.8 &	78.1 &	82.0 \\
            50\% & 72.1	& 80.4 & 75.8 & 73.7 & 83.9 &	78.5. &	82.5\\
            60\% & 72.4 & 80.6 & 76.0 & 73.9 &	84.0 &	78.6 &	82.7\\
            70\% & 72.5 & 80.5 & 76.1 &	74.1 &	83.9 & 78.7 & 82.8\\
            80\% & 72.9 & 80.7 & 76.3 &	74.3 &	84.1 & 78.9 & 83.0 \\
            90\% & 72.9 & 80.7 &  76.4	& 74.5 & 84.1 &	79.0	&83.1 \\
             \Xhline{3\arrayrulewidth} 
        \end{tabular}
        } 
    \end{center}
\end{table}

\begin{table}
    \begin{center}
     \caption{\small \textbf{Performance of MLR with partial labels on VOC2007}}~\label{table:voc_full}
        \resizebox{0.9\linewidth}{!}{
        \begin{tabular}{c c c c c c c c }
            \Xhline{3\arrayrulewidth} 
            Amount of Labels & CP & CR & CF1& OP & OR & OF1 & mAP \\
            \Xhline{3\arrayrulewidth} 
            10\% &  69.6 &	91.3 &	78.0 & 	72.4 &	92.4 &	81.2 &	90.3  \\
            20\% & 74.2 & 	92.6 &	81.7 &	76.2 &	93.6 &	84.0 & 92.2  \\
            30\% & 74.9 & 92.8 & 82.3 &	78.6 &	93.3 &	85.3 &	92.8 \\
            40\% & 78.4 & 92.5 & 84.5 & 80.8 & 93.3 & 86.6 & 93.3  \\
            50\% & 80.6 & 93.4 & 86.3 &	82.4 &	94.0 &	87.8 &	93.6 \\
            60\% & 80.1 & 93.7 & 86.0 &	81.4 &	94.4 &	87.4 &	93.9 \\
            70\% & 80.9 & 93.4 & 86.5 &	82.7 & 94.0 & 88.0 & 94.0 \\
            80\% & 80.8 & 93.8 & 86.5 &  82.9 & 94.2 & 88.2 & 94.1  \\
            90\% & 80.5	& 93.9 &	86.3 &	82.4 &	94.4 &	88.0 &	94.2 \\
             \Xhline{3\arrayrulewidth} 
        \end{tabular}
        } 
    \end{center}
\end{table}

\begin{figure}[h]
\vspace{20pt}
    \centering
    \includegraphics[width=\linewidth]{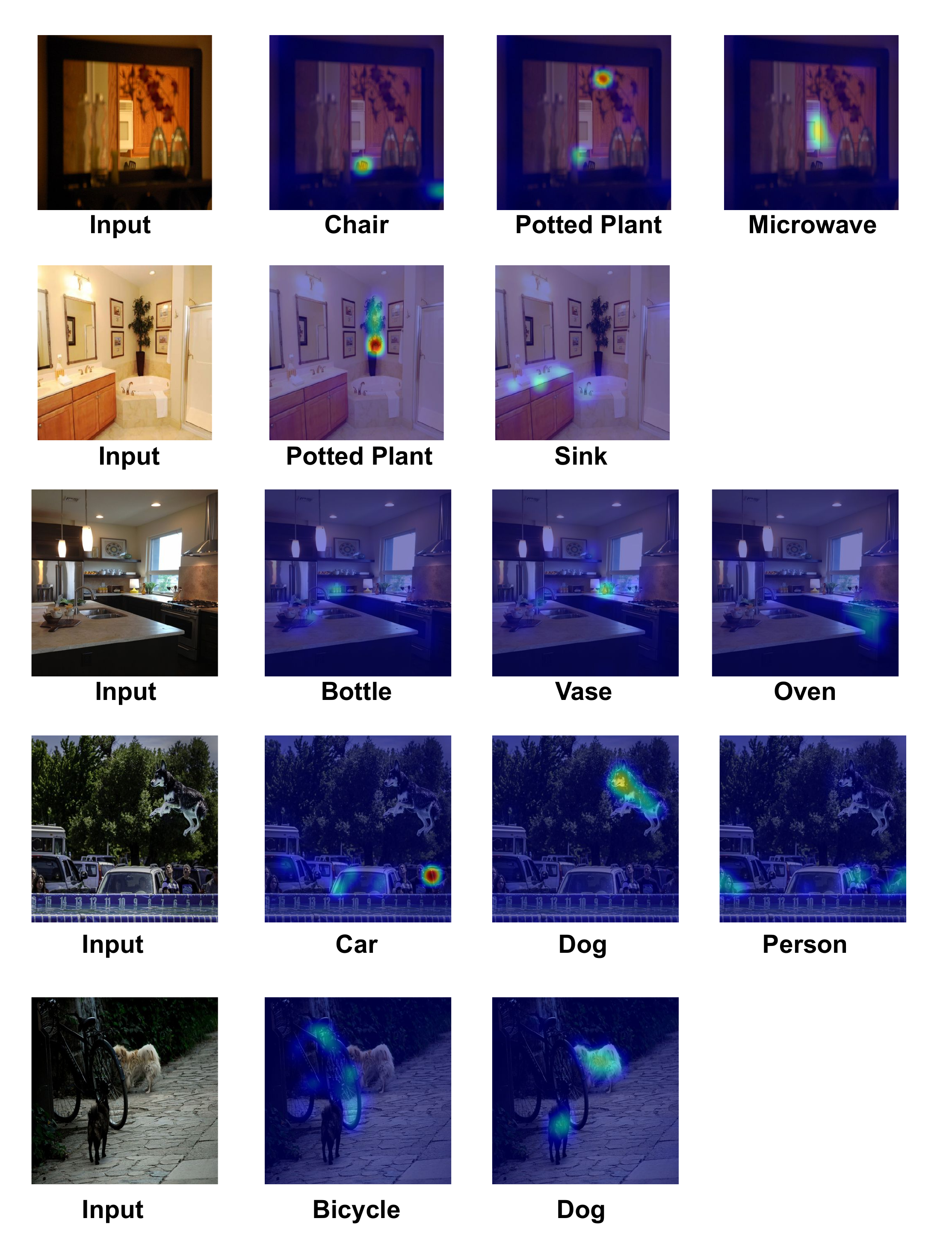} 

    \caption{\small\textbf{Visualization of Class-Specific Region Feature Aggregation} }
    \label{fig:visual} 
\end{figure}

\section{Visualization of Class-Specific Region Feature Aggregation}

We have visualized the class-specific region feature aggregation on MS-COCO dataset (in Fig.~\ref{fig:visual}). We can see \ours generates the high attention score at the correct objects.

\end{document}